\documentclass[sigconf]{acmart}

\AtBeginDocument{%
  \providecommand\BibTeX{{%
    Bib\TeX}}}

\setcopyright{acmcopyright}
\copyrightyear{2018}
\acmYear{2018}
\acmDOI{XXXXXXX.XXXXXXX}

\acmConference[Conference acronym 'XX]{Make sure to enter the correct
  conference title from your rights confirmation emai}{June 03--05,
  2018}{Woodstock, NY}
\acmPrice{15.00}
\acmISBN{978-1-4503-XXXX-X/18/06}

\usepackage{amsmath}
\usepackage{amsfonts}
\usepackage{graphicx}
\usepackage{subcaption}
\usepackage{rotating}
\usepackage{multirow}
\usepackage{pgfplots}
\pgfplotsset{compat=newest}
\usepackage{cite}
\usepackage{hyperref}
\usepackage{comment}
\usepackage{xcolor}
\usepackage{soul}
\newcommand{\conf}[1]{\scriptsize{$\pm$#1}}

\begin{document}

\title{Deep Features for CBIR with Scarce Data using Hebbian Learning}

\author{Gabriele Lagani}
\authornote{Corresponding}
\email{gabriele.lagani@phd.unipi.it}
\author{Davide Bacciu}
\email{davide.bacciu@unipi.it}
\author{Claudio Gallicchio}
\email{claudio.gallicchio@unipi.it}
\affiliation{%
  \institution{Dept. of Computer Science, University of Pisa}
  \city{Pisa}
  \country{Italy}
}

\author{Fabrizio Falchi}
\email{fabrizio.falchi@cnr.it}
\author{Claudio Gennaro}
\email{claudio.gennaro@cnr.it}
\author{Giuseppe Amato}
\email{giuseppe.amato@cnr.it}
\affiliation{%
  \institution{ISTI - CNR}
  \city{Pisa}
  \country{Italy}
}

\renewcommand{\shortauthors}{Lagani et al.}

\begin{abstract}
Features extracted from Deep Neural Networks (DNNs) have proven to be very effective in the context of Content Based Image Retrieval (CBIR).
In recent work, biologically inspired \textit{Hebbian} learning algorithms have shown promises for DNN training.
In this contribution, we study the performance of such algorithms in the development of feature extractors for CBIR tasks.
Specifically, we consider a semi-supervised learning strategy in two steps: first, an unsupervised pre-training stage is performed using Hebbian learning on the image dataset; second, the network is fine-tuned using supervised Stochastic Gradient Descent (SGD) training. For the unsupervised pre-training stage, we explore the nonlinear Hebbian Principal Component Analysis (HPCA) learning rule. For the supervised fine-tuning stage, we assume sample efficiency scenarios, in which the amount of labeled samples is just a small fraction of the whole dataset.
Our experimental analysis, conducted on the CIFAR10
and CIFAR100 datasets, 
shows that, when few labeled samples are available, our Hebbian approach provides relevant improvements compared to various alternative methods.
\end{abstract}




\maketitle

\section{Introduction}
Image retrieval is a relevant functionality used in the media sector. Consider for instance, the production of newspaper articles, where images should be retrieved to be associated with the text. In this case, an effective image retrieval engine cannot simply rely on pre-determined and pre-trained tools. It should be able to quickly intercept new emerging trends, new hot topics, and new relevant events. Accordingly, it must be constantly updated and trained with the possibly scarce and noisy data at hand. Effective solutions able to learn from scarce data allow media companies to immediately react to the occurrence of new emerging topics and be able to effectively detect media related to them, with minimum effort, minimum delay, and high accuracy. In this paper we propose the use of \textit{Hebbian} learning to build effective image feature extractors, even with scarce training data.

In the past few years, Deep Neural Networks (DNNs) have emerged as a powerful technology in the domain of computer vision \citep{krizhevsky2012, he2016}. Accordingly, the world of Content Based Image Retrieval (CBIR) observed a paradigm shift, from handcrafted features, to representations extracted from DNNs \citep{wan2014, babenko2014}.

DNN training is typically based on supervised end-to-end Stochastic Gradient Descent (SGD) algorithm with error backpropagation (\textit{backprop}). However, this approach is data-hungry, as it needs large amounts of labeled data to achieve high performances.
Moreover, the backpropagation mechanism is considered biologically implausible by neuroscientists \citep{oreilly}; instead, they suggest Hebbian learning as a candidate model for synaptic plasticity in the brain \citep{haykin, gerstner}. Notably, Hebbian approaches do not require backpropagation, nor supervision. In addition, as shown in this paper, Hebbian algorithms are quite effective when labeled data is scarce, since they can extract knowledge from the whole (possibly unlabelled) dataset, in addition to supervised training on the (scarce) labeled samples only.

In this paper, we considered a semi-supervised learning approach in two phases: first, an unsupervised pre-training phase was performed using Hebbian algorithms; after that, the DNN was fine-tuned using supervised SGD training.
Unsupervised pre-training is known to be beneficial before further fine-tuning stage \citep{bengio2007, larochelle2009, kingma2014, zhang2016}.
Concerning the Hebbian pre-training phase, a nonlinear learning rule for Hebbian Principal Component Analysis (HPCA) was considered \citep{lagani2021b, lagani2021c, lagani2022}. For the supervised fine-tuning stage, we considered sample efficiency scenarios, in which only a fraction of the training data is assumed to be labeled.

After this training process, the resulting network was used to map queries and dataset images to compact feature representations. These features were then used to perform similarity-based retrieval, evaluating the resulting precision. 
We performed experiments on CIFAR10 
and CIFAR100 \citep{cifar} datasets, 
in different sample efficiency regimes (i.e. varying the percentage of labeled samples). 
The results show that our approach outperforms alternative methods in almost all the cases, especially when the fraction of labeled samples is low.

We summarize the contributions of this paper as follows:
\begin{itemize}
    \item We introduce an approach that leverages Hebbian learning for DNN training, which is applied for the first time on CBIR tasks;
    \item We adopt a semi-supervised training approach for DNNs, in which an unsupervised Hebbian pre-training stage is followed by a supervised SGD fine-tuning stage;
    \item We provide an experimental evaluation of the resulting DNN features in CBIR tasks, considering different sample efficiency regimes.
\end{itemize}

The remainder of this paper is structured as follows:
Section~\ref{sec:rel_work} gives an overview of related work concerning applications of DNNs to CBIR, and semi-supervised learning;
Section~\ref{sec:method} illustrates our semi-supervised approach based on Hebbian learning, and its application to CBIR;
Section~\ref{sec:exp_setup} delves into the details of our experimental setup;
In Section~\ref{sec:results}, the results of our simulations are illustrated;
Finally, Section~\ref{sec:conclusions} presents our conclusions and outlines possible future developments.

\section{Related work} \label{sec:rel_work}

Deep learning technologies have demonstrated very high performance in image recognition \citep{krizhevsky2012, he2016}, despite some limitations due to vulnerability to adversarial attacks \citep{carrara2019}. In the field of CBIR, deep learning has also driven the transition from handcrafted to learned features, in the development of image descriptors for retrieval tasks \citep{amato2019deep, wan2014, babenko2014, yue2015, gordo2016, gordo2017, bai2018}.



As stated in \citep{wan2014}, the advantage of learned features over handcrafted ones lies in the reduction of the semantic gap between features: deep learning allows to obtain more abstract image descriptors that better capture the perceived similarity among objects over low-level handcrafted features. Building on this observation, past work (\citep{wan2014, babenko2014}) provided experimental evaluation of DNN features in image retrieval tasks, showing remarkable performance.

While these contributions used abstract features from higher layers, other work investigated the integration of local features into global descriptors \citep{babenko2015, yue2015, gordo2016}. In particular, features from intermediate layers were shown to better retain information about local details that are useful in the retrieval task \citep{babenko2015, yue2015}. A different approach was taken in \citep{gordo2016}, where a region proposal network was used to select regions in the input for successive processing. Features extracted from such regions were then aggregated in a global descriptor. 



In \citep{gordo2017}, the authors proposed an end-to-end training procedure in which a siamese architecture was used to process triplets of images in parallel, pushing representations of related images (according to the ground-truth) closer together, while separating those of unrelated images.



In typical CBIR scenarios, a very large dataset is given, but labels are available only for a small subset of data. The Semi-supervised Paradigm for Local Hashing (SPLH) \citep{wang2012} was proposed to exploit also the unlabeled examples. In this case, label scarcity was addressed with an information theoretic regularizer over labeled and unlabeled samples. This approach is related to pseudo-labeling and consistency-based methods in semi-supervised learning \citep{iscen2019, sellars2021}. An alternative approach for extracting information from unlabeled samples is based on unsupervised pre-training \citep{bengio2007, larochelle2009, kingma2014, zhang2016}. Our contribution falls in this second category, but we explore bio-inspired Hebbian learning for the unsupervised pre-training phase. However, pseudo-labeling/consistency-based methods and unsupervised pre-training are not in contrast with each other, and they could be integrated together. This possible future direction will also be highlighted in Section \ref{sec:conclusions}.

Finally, Bai et al. \citep{bai2018} presented a comprehensive experimental comparison of various methods on modern computer vision datasets, including their proposed Optimized AlexNet for Image Retrieval (OANIR) approach, in which they applied an AlexNet-inspired \citep{krizhevsky2012} network architecture specifically modified and optimized for the retrieval task.

\section{Semi-supervised Hebbian-SGD training for CBIR} 
\label{sec:method}

In this section we describe our contribution more in detail. We propose a  semi-supervised DNN training protocol in two phases: in the first phase, the network is pre-trained with unsupervised Hebbian learning; in the second phase, the network is fine-tuned with supervised SGD training.  After that, the resulting model is used to map images to feature representations, which can be employed in CBIR tasks. The following subsections describe each phase. Also, we face the problem of data scarcity, since, in real scenarios, for instance from the media sector, the available labeled data are typically just a small portion of the whole dataset. Hence, we describe how our unsupervised pre-training phase helps to improve the overall model by exploiting unlabeled data in addition to labeled images.

\subsection{Unsupervised Hebbian pre-training}

In the first phase of the proposed approach, the DNN model is pre-trained using approaches grounded in the neuroscientific theory of Hebbian learning \citep{gerstner}. The details of Hebbian learning theories are outside the scope of this paper, although a plethora of Hebbian-based learning algorithms have been proposed. We refer the interested reader to \citep{haykin, amato2019, lagani2021b, lagani2021c, lagani2021d, lagani2022}. In particular, in our previous works \citep{lagani2021b, lagani2021c}, we have highlighted the merits of the nonlinear Hebbian Principal Component Analysis (HPCA). 
This is derived by minimizing the \textit{representation error}:
\begin{equation}
    L(\mathbf{w_i})  = \frac{1}{2} \, E[(\mathbf{x} - \sum_{j=1}^i f(y_j) \, \mathbf{w_j})^2]
\end{equation}
Where $\mathbf{x}$ is the current input, $\mathbf{w_j}$ is the $j^{th}$ neuron weight vector, $y_j = \mathbf{w_j}^T \, \mathbf{x} $ is its output, and $f$ is the activation function. It can be shown \citep{karhunen1995, becker1996a} that such minimization leads to the following weight update rule:
\begin{equation} \label{eq:lrn_rule}
    \Delta \mathbf{w_i} = \eta f(y_i) (x - \sum_{j=1}^i f(y_j) \mathbf{w_j})
\end{equation}

Where $\eta$ is the learning rate.

Since we found this learning rule to be particularly effective in semi-supervised regimes with label scarcity \citep{lagani2021b, lagani2021c}, this is the rule that we adopted in the unsupervised pre-training phase of our method.

\subsection{Supervised SGD fine-tuning}

After the unsupervised Hebbian pre-training stage, a supervised fine-tuning phase, based on SDG with backprop takes place. This corresponds to ordinary supervised DNN training, but starting from a pre-trained network. The advantage of such a starting point lies in the fact that pre-trained neural networks are more likely to start closer to an improved local minimum \citep{bengio2007, larochelle2009}. More importantly, the two-phases semi-supervised training procedure is especially effective in scenarios with label scarcity, as described in the next subsection.

\subsection{Dealing with data scarcity}

Let's define the \textit{labeled set} $\mathcal{T}_L$ as a collection of elements for which the corresponding label is known. Conversely, the \textit{unlabeled set}  $\mathcal{T}_U$ is a collection of elements whose labels are unknown. The whole \textit{training set} $\mathcal{T}$ is given by the union of $\mathcal{T}_L$ and $\mathcal{T}_U$. All the samples from $\mathcal{T}$ are assumed to be drawn from the same statistical distribution.
Typical retrieval scenarios present a \textit{sample efficiency} issue, i.e. the number of samples in $\mathcal{T}_L$ is much smaller than the total number of samples in $\mathcal{T}$.
In particular, an $s \, \%$-sample efficiency \textit{regime} is characterized by $|\mathcal{T}_L| = \frac{s}{100}|\mathcal{T}|$ (where $|\cdot|$ denotes the cardinality of a set, i.e. the number of elements inside the set).
In other words, the size of the labeled set is $s \, \%$ that of the whole training set. 

Traditional supervised approaches based on SGD and backprop work well provided that the size of the labeled set is sufficiently large, but they do not exploit the unlabeled set. The proposed semi-supervised approach is effective to tackle this limitation. In fact, during the unsupervised phase, general information is extracted using all the available training samples, unlabeled and labeled (but without using label information in the latter case).
During the second phase, supervised fine-tuning extracts further task-specific information using only the few labeled samples (with the corresponding labels).

\subsection{Neural features for image retrieval}

Features are obtained as vectors of activations of neurons in a selected network layer, in correspondence of an input image.
Features extracted from the trained network are then used in the retrieval task as described in the following.

All dataset images are first mapped to the corresponding features, which are stored together with the associated class label. Never-seen-before test set images are used as queries. Queries are mapped to features as well. A query is executed by retrieving the most similar database images to the query image. Similarity between images is assessed measuring the euclidean distance between the corresponding features. The ground-truth to evaluate correct retrieval is defined based on the labels associated to images: a retrieved image matches the query image if they have the same label.
Note that, in this phase we are using the label information of all the data elements, but only for evaluation, and not for training.

\section{Experimental setup} \label{sec:exp_setup}

In the following, we describe the details of our experiments and comparisons, discussing the network architecture and the training procedure\footnote{Code available at: \\ \texttt{github.com/GabrieleLagani/HebbianPCA/tree/hebbretr}}.

The experiments were performed on CIFAR10, 
and CIFAR100 \citep{cifar} datasets.

The CIFAR10 dataset contains 50,000 training images and 10,000 test images, belonging to 10 classes. Moreover, the training images were randomly split into a training set of 40,000 images and a validation set of 10,000 images.

The CIFAR100 dataset also contains 50,000 training images and 10,000 test images, belonging to 100 classes. Also in this case, the training images were randomly split into a training set of 40,000 images and a validation set of 10,000 images.


We considered sample efficiency regimes in which the amount of labeled samples was respectively 1\%, 2\%, 3\%, 4\%, 5\%, 10\%, 25\%, and 100\% of the whole training set. 

\subsection{Network architecture and training}

\begin{figure*}[t]
\centering
\includegraphics[width=0.6\textwidth]{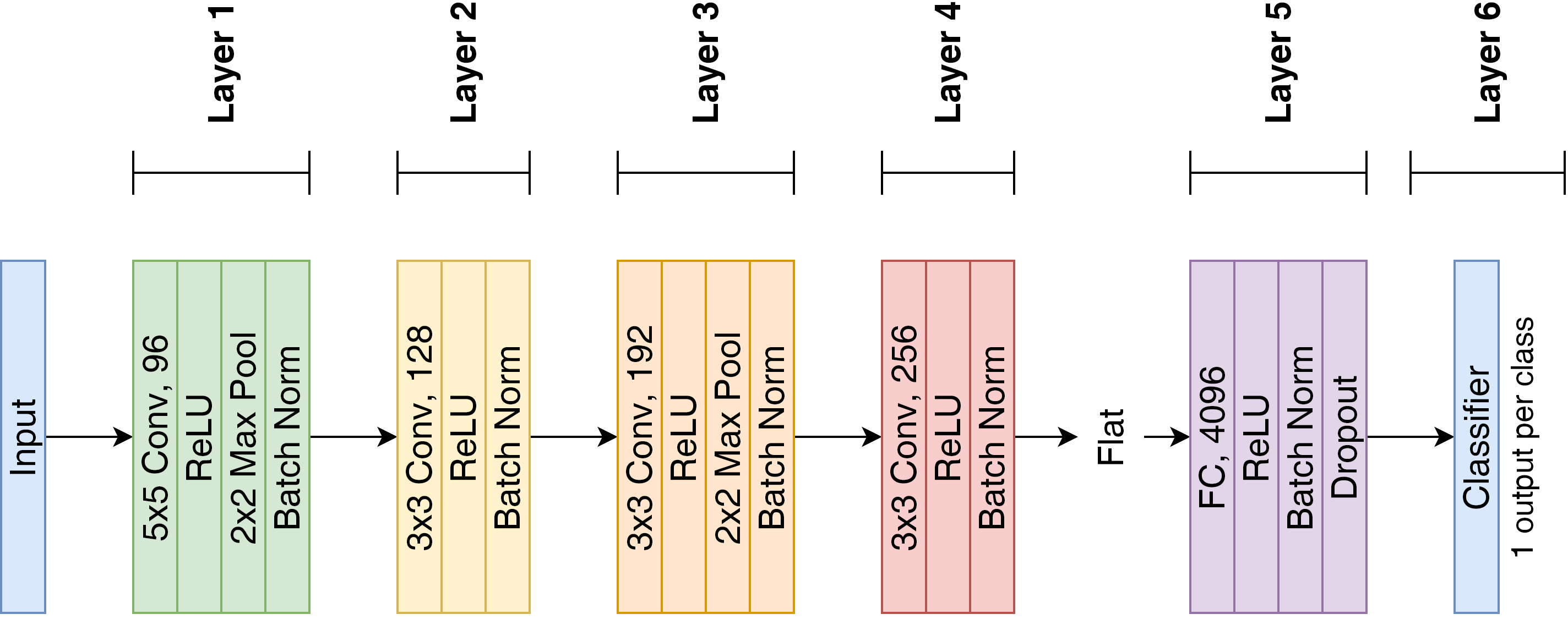}
\caption{Architecture of the DNN used in our experiments.}
\label{fig:network}
\end{figure*}

We considered a six layer neural network as shown in Fig. \ref{fig:network}: five deep layers plus a final linear classifier. The various layers were interleaved with other processing stages (such as ReLU nonlinearities, max-pooling, etc.). The architecture was inspired by AlexNet \citep{krizhevsky2012}, but with slight modifications in order to reduce the overall computational cost of training. 
We decided to adopt a simple network model in order to be able to evaluate the effects of different learning approaches on a layer-by-layer basis. This choice also makes it more practical for other researchers to reproduce the experiments.

For each sample efficiency regime, we trained the network with our semi-supervised approach in a classification task. First, we used HPCA unsupervised pre-training (Eq. \ref{eq:lrn_rule}) in the internal layers. This was followed by the fine tuning stage with SGD training, involving the final classifier as well as the previous layers, in an end-to-end fashion.

For each configuration we also created a baseline for comparison. In this case, we used another popular unsupervised method, namely the Variational Auto-Encoder (VAE) \citep{kingma2013}, for the unsupervised pre-training stage. This was again followed by the supervised end-to-end fine tuning based on SGD. VAE-based semi-supervised learning was also the approach considered in \citep{kingma2014}. We also considered the case with no pre-training.

\subsection{Layerwise evaluation of network features for CBIR}

The resulting trained models were evaluated in the CBIR task. For this purpose, trained networks were used to extract feature representations from dataset and query images. These features were then used to rank dataset images by similarity with the query, evaluating the retrieval performance.

As an evaluation metric, we considered the Average Precision Score (APS) 
:
\begin{equation}
    APS = \sum_{i=1}^K P_i \, (R_i - R_{i-1})
\end{equation}
where $P_i$ is the precision at the $i^{th}$ retrieved item, $R_i$ is the corresponding recall. This score is 
averaged over all the queries, thus obtaining the mean Average Precision (mAP).

In our experiments, we also evaluated the quality of the features from internal layers in the CBIR task, in order to find the layer that provided the best performance. For this purpose, we cut the networks in correspondence of a given layer, and we placed a new classifier on top of it. The supervised fine-tuning stage was then performed, in an end-to-end fashion, on the classifier and the preceding layers, and the resulting retrieval precision was evaluated. This was done for each layer, one at a time, in order to evaluate the quality of the neural features on a layer by layer basis. Evaluation for the selection of the best layer was performed on a validation set of images that are distinct from the training images. After that, the final evaluation was performed on a test set of never-seen-before images, and the results are shown in Section \ref{sec:results}.

\subsection{Details of training}

We implemented our experiments using PyTorch. All the hyperparameters mentioned below resulted from a parameter search aimed at maximizing the validation accuracy on the respective datasets, following the Coordinate Descent (CD) approach \citep{kolda2003}. 

Network weights were initialized following \textit{Xavier} initialization \citep{glorot2010}, which in some cases performs better than pre-training.

Training was performed in 20 epochs using mini-batches of size 64. No more epochs were necessary, since the models had already reached convergence at that point. Networks were fed input images of size 32x32 pixels.

In the Hebbian training, the learning rate was set to $10^{-3}$. No L2 regularization or dropout was used, since the learning method did not present overfitting issues.

For VAE training, the network in Fig. \ref{fig:network}, up to layer 5, acted as encoder, with an extra layer mapping layer 5 output to 256 gaussian latent variables, while a specular network branch acted as decoder.
VAE training was performed without supervision, in an end-to-end encoding-decoding task, optimizing the $\beta$-VAE Variational Lower Bound \citep{higgins2016}, with coefficient $\beta = 0.5$

For the supervised training stage, based on SGD, the initial learning rate was set to $10^{-3}$ and kept constant for the first ten epochs, while it was halved every two epochs for the remaining ten epochs. We also used momentum coefficient $0.9$, Nesterov correction, dropout rate 0.5 and L2 weight decay penalty coefficient set to $5 \cdot  10^{-2}$ for CIFAR10, 
and $10^{-2}$ for CIFAR100. 
Cross-entropy loss was used as optimization metric.

To obtain the best possible generalization, \textit{early stopping} was used in each training session, i.e. we chose as final trained model the state of the network at the epoch when the highest validation accuracy was recorded.

\section{Results and discussion} \label{sec:results}

In this section, the experimental results obtained with each dataset are presented and analyzed. We report the classification accuracy, along with 95\% confidence intervals obtained from five independent iterations of the experiments, in the various sample efficiency regimes, for the CIFAR10, 
and CIFAR100 datasets.

\subsection{CIFAR10}

\begin{table}[t]
	\caption{CIFAR10 mAP test results and corresponding layer used for feature extraction.} 
	\begin{center} 
		\begin{tabular}{cccc} 
			\hline 
			\textbf{Regime} & \textbf{Pre-train} & \textbf{mAP (\%)} & \textbf{Layer} \\ 
			\hline 
			\multirow{3}{*}{1\%} 
				& None & 14.17 \conf{0.04} & 3 \\ 
				& VAE & 13.79 \conf{0.04} & 1 \\ 
				& HPCA & \textbf{17.81} \conf{0.04} & 3 \\ 
			\hline 
			\multirow{3}{*}{2\%} 
				& None & 14.70 \conf{0.04} & 3 \\ 
				& VAE & 14.39 \conf{0.05} & 1 \\ 
				& HPCA & \textbf{18.33} \conf{0.05} & 3 \\ 
			\hline 
			\multirow{3}{*}{3\%} 
				& None & 15.65 \conf{0.05} & 5 \\ 
				& VAE & 14.79 \conf{0.06} & 1 \\ 
				& HPCA & \textbf{18.84} \conf{0.05} & 3 \\ 
			\hline 
			\multirow{3}{*}{4\%} 
				& None & 17.03 \conf{0.06} & 5 \\ 
				& VAE & 15.49 \conf{0.05} & 2 \\ 
				& HPCA & \textbf{19.43} \conf{0.04} & 3 \\ 
			\hline 
			\multirow{3}{*}{5\%} 
				& None & 18.61 \conf{0.03} & 5 \\ 
				& VAE & 16.19 \conf{0.07} & 2 \\ 
				& HPCA & \textbf{20.08} \conf{0.05} & 3 \\ 
			\hline 
			\multirow{3}{*}{10\%} 
				& None & \textbf{25.48} \conf{0.07} & 5 \\ 
				& VAE & 21.58 \conf{0.05} & 3 \\ 
				& HPCA & 22.54 \conf{0.02} & 3 \\ 
			\hline 
			\multirow{3}{*}{25\%} 
				& None & \textbf{42.18} \conf{0.07} & 5 \\ 
				& VAE & 40.75 \conf{0.07} & 5 \\ 
				& HPCA & 27.29 \conf{0.03} & 3 \\ 
			\hline 
			\multirow{3}{*}{100\%} 
				& None & \textbf{81.26} \conf{0.01} & 5 \\ 
				& VAE & 80.56 \conf{0.01} & 5 \\ 
				& HPCA & 80.01 \conf{0.01} & 5 \\ 
			\hline 
		\end{tabular} 
 		\label{tab:cifar10} 
 	\end{center} 
\end{table} 

Tab. \ref{tab:cifar10} reports the mAP results obtained on the CIFAR10 dataset.

As we can observe, in regimes of extreme sample efficiency (below 10\%), where the number of available labeled samples is very scarce, HPCA always performs better than VAE and no pre-training. In particular, in the 1\% sample efficiency case, HPCA outperforms VAE pre-training by about 4\%. Note that HPCA pre-training, with only 5\% regime, obtains a mAP equal to 1/4 of that obtained with 100\% regime.
VAE pre-training starts to perform better only when the number of labeled samples increases, although no additional benefit is achieved in these cases compared to no pre-training. This is in contrast with what happens in image classification tasks, where previous results showed the effectiveness of VAE pre-training in such sample efficiency regimes \citep{lagani2021b}. 
However, it is known that appropriate weight initialization, such as the one that we have used for our networks, can be more effective than pre-training in some cases \citep{glorot2010}.

In the Table, the Layer column reports the layer used as feature extractor. This layer was selected as the one producing the best performance during validation. Notably, test experiments conducted with features extracted from all the layers confirmed that the best choice taken during validation actually corresponds to the layer offering the best performance on the test set.

\subsection{CIFAR100}

\begin{table} 
	\caption{CIFAR100 mAP test results and corresponding layer used for feature extraction.} 
	\begin{center} 
		\begin{tabular}{cccc} 
			\hline 
			\textbf{Regime} & \textbf{Method} & \textbf{mAP (\%)} & \textbf{Layer} \\ 
			\hline 
			\multirow{3}{*}{1\%} 
				& None & 2.48 \conf{0.02} & 1 \\ 
				& VAE & 2.37 \conf{0.02} & 1 \\ 
				& HPCA & \textbf{2.95} \conf{0.03} & 3 \\ 
			\hline 
			\multirow{3}{*}{2\%} 
				& None & 2.54 \conf{0.02} & 3 \\ 
				& VAE & 2.5 \conf{0.02} & 1 \\ 
				& HPCA & \textbf{3.1} \conf{0.03} & 3 \\ 
			\hline 
			\multirow{3}{*}{3\%} 
				& None & 2.61 \conf{0.02} & 3 \\ 
				& VAE & 2.59 \conf{0.02} & 1 \\ 
				& HPCA & \textbf{3.18} \conf{0.03} & 3 \\ 
			\hline 
			\multirow{3}{*}{4\%} 
				& None & 2.64 \conf{0.02} & 5 \\ 
				& VAE & 2.64 \conf{0.02} & 1 \\ 
				& HPCA & \textbf{3.28} \conf{0.03} & 3 \\ 
			\hline 
			\multirow{3}{*}{5\%} 
				& None & 2.82 \conf{0.03} & 5 \\ 
				& VAE & 2.68 \conf{0.02} & 1 \\ 
				& HPCA & \textbf{3.34} \conf{0.03} & 3 \\ 
			\hline 
			\multirow{3}{*}{10\%} 
				& None & \textbf{3.63} \conf{0.02} & 5 \\ 
				& VAE & 3.09 \conf{0.03} & 2 \\ 
				& HPCA & 3.57 \conf{0.03} & 3 \\ 
			\hline 
			\multirow{3}{*}{25\%} 
				& None & \textbf{5.94} \conf{0.03} & 5 \\ 
				& VAE & 4.71 \conf{0.04} & 3 \\ 
				& HPCA & 4.27 \conf{0.02} & 3 \\ 
			\hline 
			\multirow{3}{*}{100\%} 
				& None & \textbf{18.86} \conf{0.02} & 5 \\ 
				& VAE & 17.68 \conf{0.04} & 5 \\ 
				& HPCA & 8.41 \conf{0.03} & 5 \\ 
			\hline 
		\end{tabular} 
 		\label{tab:cifar100} 
 	\end{center} 
\end{table}

Since CIFAR10 contained just 10 different classes, to validate our observations with a similar, yet more difficult scenario, we also performed tests with CIFAR100, containing 100 classes.
Tab. \ref{tab:cifar100} reports the mAP results obtained on the CIFAR100 dataset. 

These results confirm the previous observations. In sample efficiency regimes below 10\%, HPCA performs better than VAE pre-training and no pre-training. The gap w.r.t. the previous case is reduced, but it still remains statistically significant. In particular, in the 5\% sample efficiency case, HPCA outperforms VAE pre-training by almost 1 percent point. Again, VAE pre-training starts to perform better only when the number of labeled samples increases, although no additional benefit is achieved in these cases compared to no pre-training.
On the other hand, HPCA pre-training seems to be detrimental when a larger number of samples is available, as it could move the weights towards a configuration where a good local optimum is harder to reach in the successive fine-tuning.

Note that CIFAR 100 represents a more difficult benchmark for the retrieval task, with respect to CIFAR 10.  
In fact, a low mAP is achieved even in the 100\% sample efficiency regime.

\section{Conclusions and future work}
\label{sec:conclusions}
In this paper we have explored a semi-supervised training strategy for DNNs based on biologically inspired Hebbian learning, applied for the first time in the context of CBIR. We focused on scenarios of scarce data availability, which is relevant, for example, in the media sector.

Our experimental analysis, conducted on different datasets, suggest that our semi-supervised approach based on the unsupervised HPCA algorithm performs generally better than VAE pre-training or no pre-training, especially in sample efficiency regimes of extreme label scarcity, in which only a small portion of the training set (below 10\%) is assumed to be labeled. 
On the other hand, VAE pre-training seems to become more effective in regimes where a larger portion of the training set (above 10\%) is labeled. 
Therefore, our method is preferable in scenarios in which manually labeling a large number of training samples would be too expensive, while gathering unlabeled samples is relatively cheap.

In future works, we plan to provide a more thorough experimental validation of the proposed method also on more complex datasets.
Moreover, further improvements might come from exploring more complex feature extraction strategies, which can also be formulated as Hebbian learning variants, such as Independent Component Analysis (ICA) \citep{hyvarinen} and sparse coding \citep{olshausen1996a, olshausen1996b, rozell2008}, as well as from more biologically faithful neural models such as Spiking Neural Networks (SNNs) \citep{gerstner, lagani2021a}. 
Finally, Hebbian approaches can also be combined with pseudo-labeling and consistency methods mentioned in Section \ref{sec:rel_work} \citep{iscen2019, sellars2021}.

\begin{acks}
This work was partially supported by the H2020 project AI4Media (GA 951911).
\end{acks}

\bibliographystyle{ACM-Reference-Format}
\bibliography{references}


\end{document}